\documentclass{article}
\usepackage{spconf,amsmath,graphicx}

\usepackage{enumitem}
\usepackage{booktabs}
\usepackage{multirow}
\usepackage{amsmath} 
\usepackage[table]{xcolor}
\usepackage{makecell} 
\usepackage{siunitx} 
\usepackage{pifont}
\usepackage{float}
\usepackage{dirtytalk}
\usepackage{stfloats}
\usepackage{url}

\newcommand{\cmark}{\ding{51}} 
\newcommand{\xmark}{\ding{55}} 
\newcommand{\cint}[2]{\renewcommand\cellset{\renewcommand\arraystretch{0.5}}\makecell[tc]{#1\\\textcolor[gray]{0.3}{\tiny[#2]}}}
\setlist{nosep, leftmargin=14pt}

\usepackage{mwe} 

\title{LungEvaty: A Scalable, Open-Source Transformer-based Deep Learning Model
for Lung Cancer Risk Prediction in LDCT Screening}

\name{%
  \begin{tabular}{@{}c@{}}
    Johannes Brandt$^{1,\star}$\sthanks{$^\star$~Shared first authorship. $\dagger$~Equal supervision.}, 
    Maulik Chevli$^{1,\star}$, Rickmer Braren$^{1,4,5}$,\\
    {\itshape Georgios Kaissis$^{1}$, Philip M\"uller$^{1,\dagger}$, and Daniel Rueckert$^{1,2,3,\dagger}$}
  \end{tabular}
}

\address{$^{1}$ Chair for AI in Healthcare and Medicine, Technical University of Munich (TUM) \\
and TUM University Hospital, Munich, Germany \\
$^{2}$ Dept. of Computing, Imperial College London, UK \\
$^{3}$ Munich Center for Machine Learning (MCML), Munich, Germany \\
$^{4}$ Institute of Diagnostic and Interventional Radiology, TUM, School of Medicine, Munich, Germany \\
$^{5}$ Dept. of Diagnostic and Interventional Radiology, UKE Hamburg, Germany
}

\begin{document}
\maketitle

\begin{abstract}
Lung cancer risk estimation is gaining increasing importance as more countries introduce population-wide screening programs using low-dose CT (LDCT). As imaging volumes grow, scalable methods that can process entire lung volumes efficiently are essential to tap into the full potential of these large screening datasets. Existing approaches either over-rely on pixel-level annotations, limiting scalability, or analyze the lung in fragments, weakening performance. We present \textbf{LungEvaty}, a fully Transformer-based framework for predicting 1–6 year lung cancer risk from a single LDCT scan. The model operates on whole-lung inputs, learning directly from large-scale screening data to capture comprehensive anatomical and pathological cues relevant for malignancy risk. Using only imaging data and no region supervision, \textbf{LungEvaty} matches state-of-the-art performance, refinable by our optional Anatomically Informed Attention Guidance (AIAG) loss, which encourages anatomically focused attention. In total, \textbf{LungEvaty} was trained on more than 90,000 CT scans, including over 28,000 for fine-tuning and 6,000 for evaluation. Our framework offers a simple, data-efficient, and fully open-source solution that provides an extensible foundation for future research in longitudinal and multimodal lung cancer risk prediction.
\end{abstract}


\section{Introduction}
\label{sec:intro}

The early detection of lung cancer through low-dose computed tomography (LDCT) screening has been shown to significantly reduce mortality~\cite{Ardila2019EndtoendLC,mikhael2023sybil}. Deep learning has emerged as a powerful tool to augment the process, with models capable of predicting a patient’s long-term cancer risk from a single scan~\cite{Ardila2019EndtoendLC,mikhael2023sybil,niu2025medical}.
Early work by Ardila et al.~\cite{Ardila2019EndtoendLC} and Mikhael et al.~\cite{mikhael2023sybil} employed CNN-based architectures that leveraged expert-annotated malignancy localizations for region extraction or region guidance. These localizations offer strong additional supervision but are limited by their high annotation cost and lack of scalability, as they depend on specialized expertise.
More recently, the generalist transformer M3FM~\cite{niu2025medical} achieved strong results on chest CT tasks without pixel-level supervision. However, their architecture has several limitations: Their model cannot leverage pixel-level annotations even when they are available, missing an opportunity for strong spatial guidance. To compensate, M3FM depends on scaling to very large model sizes and requires large amounts of multimodal, multi-task data for fine-tuning. Additionally, their architecture processes lung regions in isolation. These fragmented representations lack a holistic anatomical context required for longitudinal studies and modern multimodal pretraining, like contrastive or generative methods. Finally, the lack of publicly available code or model weights for M3FM hinders reproducibility and benchmarking.

We address these limitations with LungEvaty, a lung-specific Transformer framework for six-year cancer risk prediction from a single LDCT scan. 
It reaches state-of-the-art performance using a single modality, less data, and no expert-annotated malignancy localizations.
Our design is scalable to large cohorts that have readily available time-to-event labels, but at the same time retains the flexibility to seamlessly integrate optional pixel-level annotations for enhanced spatial supervision when available.
By processing the entire lung volume, the model learns a latent representation that captures the full anatomical context, enabling lung-specific transfer learning, longitudinal modeling, and multimodal pretraining. Our contributions are as follows:
\begin{enumerate}
    \item We introduce \textbf{LungEvaty}, a fully \textbf{Transformer-based} architecture for six-year-long lung cancer risk prediction from a single LDCT scan.
    \item We introduce \textbf{Anatomically Informed Attention Guidance} (AIAG) loss that provides regional guidance for transformer-based lung cancer risk prediction.
    \item Rooted in \textit{modern deep learning} principles, we propose a \textbf{scalable, flexible} training paradigm that works without pixel-level annotations and can optionally integrate spatial guidance for stronger supervision.
    \item We demonstrate \textit{on par or better} performance compared to the multimodal, multitask trained SOTA baseline using \textit{60\% of the pre training data volume}, a \textit{single modality}, and a \textit{single task}.
\end{enumerate}

\begin{figure*}[!t]

  \centering
  \centerline{\includegraphics[width=0.70\linewidth, trim={0 1 0 10}, clip]{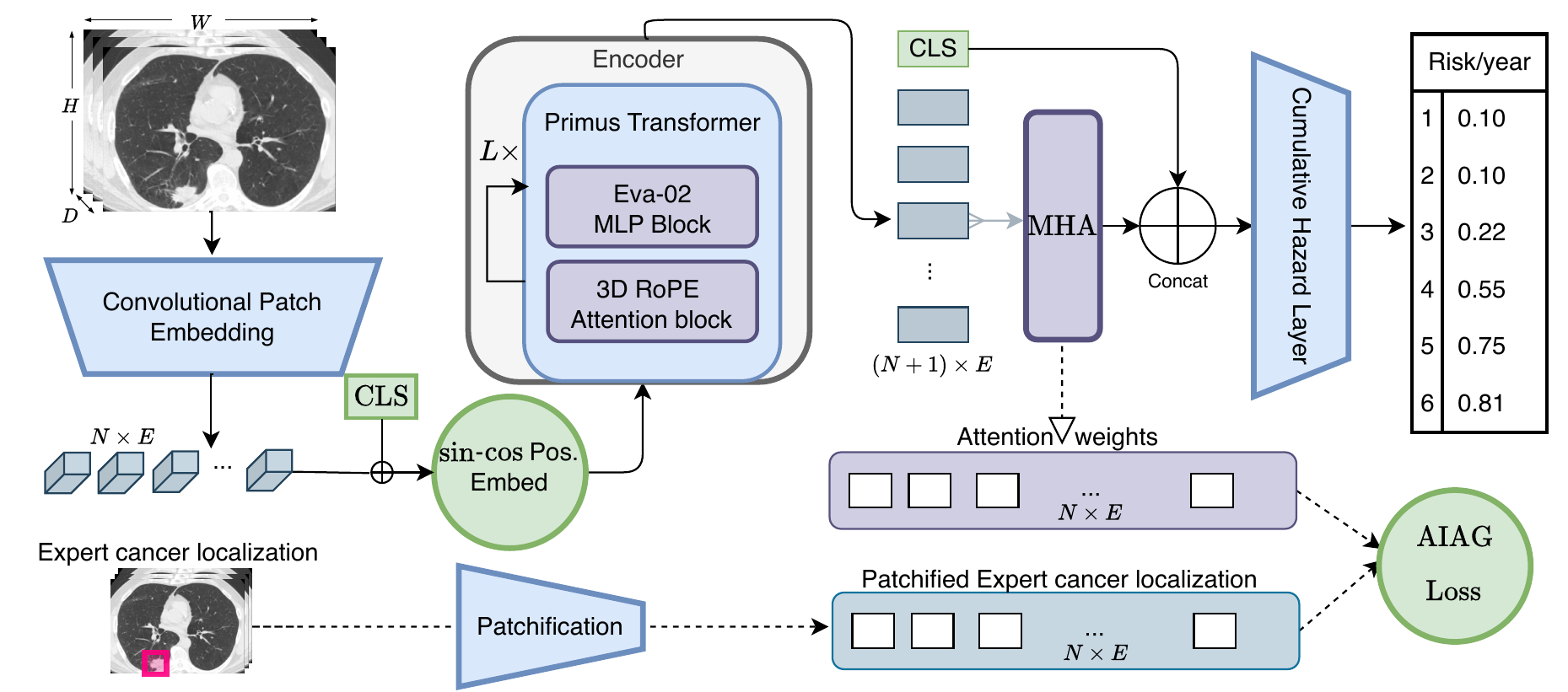}}
\caption{\textbf{LungEvaty architecture.} The encoder consists of Primus (EVA-02–style) Transformer blocks~\cite{wald2025primus}, pretrained with masked autoencoding on NLST~\cite{national2011national}. Our framework's whole-lung processing provides a rich, holistic context in the latent representation. To capture both global lung characteristics and local nodule information from this input, we use two learnable tokens: a \texttt{CLS} token and a multi-head attention query. This architecture is highly scalable for risk prediction while also supporting flexible region guidance via our novel Anatomically Informed Attention Guidance (AIAG) loss, which can leverage optional pixel-level expert annotations.}
  \label{fig: architecture}

\end{figure*}

\section{Method}
\label{sec:format}
\noindent\textbf{Overview.}
Image-based lung cancer risk estimation involves assessing the overall lung tissue, plus the identification of the smallest lesions that  may indicate malignancy.
We design our Transformer-based architecture to explicitly address this challenge by including two learnable components for global and local assessment.
As shown in Figure \ref{fig: architecture}, our model consists of a Transformer backbone that processes the whole lung, along with a learnable \texttt{CLS} token to capture holistic, global features.  
To address local malignancy, we introduce a learnable token that is used to query the subsequent multi-head attention layer, to generate an attention-pooled token focused on relevant regional features.
The \texttt{CLS} token and the attention-pooled token are concatenated and fed into a Cumulative Hazard Layer \cite{mikhael2023sybil}, which consists of linear layers with non-linear activations for final risk prediction. 
The model is optimized using a  risk loss and optionally our defined \textit{Anatomically Informed Attention Guidance (AIAG)} loss.
Our training strategy has three modules: i) Data preprocessing to remove non-lung information, ii) pre-training with masked image modeling to obtain meaningful, rich model initialization, and iii) fine-tuning to predict a patient's future lung cancer risk based on a single LDCT scan.

\noindent\textbf{Localisation of Anatomical Regions.} 
\label{sec:localisation_module}
In the data preprocessing module, 
we segment the lungs into five anatomical lobes using a segmentation model from \cite{hofmanninger2020automatic}.
We pad these segmentation masks to account for minor errors and crop the input LDCT.
This preserves lung tissue details while reducing input size and therefore the sequence length for our Transformer encoder. Additionally, anatomical lobe information can be used in the AIAG loss, as will be described later.
\\
\noindent\textbf{Masked Image Modelling (MIM).}
We pretrain our transformer backbone using self-supervised masked image modeling, following the approach outlined by He et al.~\cite{He2022}. 
We employ an asymmetric encoder-decoder architecture for this task. Instead of the standard Vision Transformer (ViT), we utilize the Eva-based Primus transformer architecture \cite{wald2025primus} as both the encoder and decoder. 
To pre-train the model, we use the NLST \cite{national2011national} training dataset, applying a Masked Autoencoder (MAE) framework. 
For reconstruction loss, we compute mean squared error on the reconstructed patches. The encoder is then used as the backbone for the downstream task.
\\
\noindent\textbf{Fine tuning.} 
\label{sec:risk_prediction_module}
Our primary objective is to predict the cumulative probability of developing cancer over a 6-year period using the NLST dataset.
We adopt the  survival modeling framework of \cite{mikhael2023sybil}, where time is discretized into yearly intervals and risk is predicted at each year, conditional on prior survival. This approach treats each screening timepoint (CT scan) independently and does not consider temporal dependencies.
Our two training objectives are described below.

\textit{1. Risk loss}:
Risk loss or  survival loss, is defined according to \cite{mikhael2023sybil} as the mean cross-entropy between the predicted risk scores 
and the ground-truth cancer outcomes \(y_n\) for each timepoint \(n \in \{1,\dots,6\}\).
If a datapoint is not censored, the label \( y_n \in \{0, 1\} \) is assigned a value of 1 (positive) if the time-to-event, relative to the current screening timepoint, is less than or equal to year \( n \); otherwise, it is assigned 0 (negative). In the case of censoring, the label \( y_n \) is always set to 0.
To maintain the isotonicity constraint of yearly risk predictions 
we apply a cumulative hazard layer \cite{yala2021toward} after the encoder and the MHA layer, as shown in Figure~\ref{fig: architecture}.

\textit{2. Anatomically Informed Attention Guidance (AIAG):} 
The final image representation combines a global \texttt{CLS} token with an attention-pooled token, obtained through a learnable query attending over all encoded embeddings. 
To encourage attention on anatomically meaningful regions, we introduce an auxiliary loss termed \textit{Anatomically Informed Attention Guidance} (AIAG), inspired by \textit{Sybil} \cite{mikhael2023sybil}. 
AIAG endorses higher attention weights in the MHA layer on image regions corresponding to known malignancy locations, using publicly available nodule bounding boxes \cite{mikhael2023sybil} and automatically generated lung lobe segmentation masks. 
Following~\cite{mikhael2023sybil}, we employ a Kullback--Leibler divergence term to align the normalized attention weights with binary nodule masks. 
In addition, we employ a region-level cross-entropy term that enforces consistency between aggregated attention and the annotated malignant lobe or, when unavailable, lung-side label. 
This auxiliary objective is applied during fine-tuning, and we systematically assess its effect on downstream risk prediction.

\section{Experiments and Results}
\label{sec:pagestyle}
\noindent\textbf{Dataset.} For pretraining, we use 91{,}827 LDCT series from 19{,}232 patients in the NLST dataset, disjoint from the validation and test splits defined by Ardila et al. For downstream risk prediction, we adopt the NLST cohort with the original train–validation–test split of Ardila et al.~\cite{Ardila2019EndtoendLC} and Mikhael et al.~\cite{mikhael2023sybil}. All CT scans are resampled to $1.4\times1.4\times2.5~\mathrm{mm}^3$, windowed to $[-1350,150]$ HU, and normalized to $[-1,1]$. 

\noindent\textbf{Implementation Details.} 
As shown in Figure \ref{fig: architecture}, our encoder is Primus \cite{wald2025primus}, a modern medical vision transformer that enforces reliance on self-attention for local and long-range dependencies while ensuring stable training on comparatively small medical datasets.
Core refinements compared to the original ViT \cite{Dosovitskiy2020} include dual positional encoding (learnable absolute embeddings combined with 3D Rotary Position Embeddings, RoPE) and EVA-02–style \cite{fang2024eva} block. 

For MIM we employ the Primus-B/8 backbone, with an encoder embedding dimension of 792, depth of 12 layers, and 12 attention heads. 
The decoder uses an embedding dimension of 396 with 4 layers and 6 attention heads. 
All experiments were trained with the AdamW optimizer using a patch size of $8 \times 8 \times 8$. 
Pretraining was conducted for 120 epochs at a constant learning rate of $3.0 \times 10^{-4}$.
During fine-tuning, we use the backbone as depicted in Figure \ref{fig: architecture} and in some experiments also optimize \textit{AIAG}. 
Fine-tuning was performed for 10 epochs: during the first 2 epochs, the encoder was frozen and the learning rate was linearly increased to $5.0 \times 10^{-5}$ for the first 10\% of those two epochs before annealing back to $5.0 \times 10^{-10}$. 
This warm-up/annealing cycle was then repeated over the remaining 8 epochs with the encoder unfrozen.

\noindent\textbf{Baselines.}
To the best of our knowledge, there are only two models that perform the same task:
\begin{enumerate}
    \item \textbf{Sybil} \cite{mikhael2023sybil}, a ResNet3D architecture pre-trained on Kinetics \cite{Kay2017} and fine-tuned on NLST data and
    \item \textbf{M3FM} \cite{niu2025medical}, a multitask text-image foundation model unimodally pretrained on NLST and MIDRC \cite{MIDRC} images.
\end{enumerate}

\noindent\textbf{Experiments.}
We optimize the model with a \textit{risk loss} and evaluate variants that add an \textit{AIAG} loss in combinations of expert annotations and lobe/side segmentations as shown in Table~\ref{tab:auc}.
For Sybil, we reimplement their training pipeline to work with our pre-processing and experiment with the same supervision regimes as above for direct comparison.
To benchmark against M3FM \cite{niu2025medical}, which reported on single-modality, single-task (SM-ST) and multi-modal, multi-task (MM-MT) \say{Huge} models using a \textit{non-standard NLST test split}, we perform a dual-split cross-evaluation.
First, we test our model and Sybil \cite{mikhael2023sybil} on the M3FM split (our replication: 2,279 patients, 10,300 series vs. their 10,308 due to processing failures). Second, we evaluate their public \say{Large} MM-MT model using our pre-processing on our standard test set (2,202 patients, 6,273 series), which aligns with \cite{mikhael2023sybil}. This serves as a robust SOTA proxy, as the M3FM authors report near-par performance for their \textit{Large} and \textit{Huge} models \cite{niu2025medical}—and is the only way for PR-AUC comparison, as they do not report it.

\noindent\textbf{Results.}
Tables \ref{tab:auc} and \ref{tab:prauc} demonstrate LungEvaty's state-of-the-art performance.
On the standard Sybil split (Table \ref{tab:auc}), our model, even without pixel-level supervision, decisively outperforms both Sybil and the M3FM-Large model across all 6-year ROC-AUC horizons and C-Index. 
On the M3FM split, which allows for a direct comparison, our single-modality model is highly competitive with M3FM's \say{Huge} MM-MT model, even surpassing it on the overall C-Index. 
This SOTA performance is achieved despite using only a single modality and a simpler SM-ST training setup.
More critically, the model's superiority is most pronounced in the PR-AUC results (Table \ref{tab:prauc}). 
Given the severe class imbalance in cancer screening, PR-AUC is a far more stringent and clinically relevant metric. 
On this measure, LungEvaty outperforms all benchmarks across every time horizon, highlighting practical utility for this task.

\begin{table*}[!t]
\centering
\footnotesize
\renewcommand{\arraystretch}{1.1} 
\caption{
\textbf{ROC-AUC scores} (Years 1-6) and overall \textbf{C-Index} (95\% CI) on the Sybil and M3FM test splits. LungEvaty demonstrates SOTA performance. For LungEvaty, pixel-level guidance (Expert Anno.) provides a modest boost for \textbf{short-term risk (Y1-Y4)}, which aligns with the strong predictive power of visible features for imminent cancer. Critically, our AIAG-free variant (gray) remains highly competitive and excels at \textbf{long-term risk (Y5-Y6)}, demonstrating that state-of-the-art risk prediction is achievable with a single modality and without pixel-level expert annotations.
}
\label{tab:auc}
\begin{tabular}{l l *{2}{c} *{6}{c} c}
\toprule
\multirow{2}{*}{\textbf{Model}} & \multirow{2}{*}{\textbf{Training}} & \multicolumn{2}{c}{\textbf{Additional Supervision}} & \multicolumn{6}{c}{\textbf{AUC (Years)}} & \multirow{2}{*}{\textbf{C-Index}} \\
\cmidrule(lr){3-4} \cmidrule(lr){5-10}
& & \textbf{Expert Anno.} & \textbf{Lung Seg.} & Y1 & Y2 & Y3 & Y4 & Y5 & Y6 & \\
\midrule
\rowcolor{blue!15}
\multicolumn{11}{c}{\textbf{\textsf{Sybil Test Split}}} \\
\midrule

\multirow{3}{*}{Sybil}  & \multirow{3}{*}{SM-ST}
& \cmark & \cmark &
    \cint{0.927}{0.893, 0.952} & \cint{0.844}{0.804, 0.882} & \cint{0.792}{0.755, 0.827} & \cint{0.769}{0.735, 0.803} & \cint{0.755}{0.722, 0.785} & \cint{0.750}{0.719, 0.779} & \cint{0.749}{0.719, 0.778} \\
& & \xmark & \cmark &
    \cint{0.789}{0.725, 0.844} & \cint{0.742}{0.697, 0.793} & \cint{0.711}{0.671, 0.754} & \cint{0.700}{0.664, 0.738} & \cint{0.687}{0.654, 0.722} & \cint{0.683}{0.651, 0.716} & \cint{0.681}{0.650, 0.714} \\
& & \xmark & \xmark &
    \cint{0.597}{0.539, 0.662} & \cint{0.624}{0.582, 0.670} & \cint{0.618}{0.578, 0.655} & \cint{0.624}{0.588, 0.657} & \cint{0.625}{0.593, 0.655} & \cint{0.630}{0.598, 0.659} & \cint{0.621}{0.590, 0.650} \\
\midrule
\multirow{1}{*}{M3FM} & Large, MM-MT
& \xmark & \xmark &
    \cint{0.892}{0.852, 0.925} & \cint{0.845}{0.811, 0.875} & \cint{0.803}{0.773, 0.830} & \cint{0.782}{0.754, 0.809} & \cint{0.770}{0.742, 0.797} & \cint{0.769}{0.742, 0.794} & \cint{0.762}{0.736, 0.787} \\
\midrule
\multirow{3}{*}{\makecell{LungEvaty \\ (Ours)}} & \multirow{3}{*}{SM-ST}
& \cmark & \cmark &
    \cint{\textbf{0.928}}{0.892, 0.959} & \cint{\textbf{0.893}}{0.860, 0.920} & \cint{\textbf{0.848}}{0.814, 0.876} & \cint{\underline{0.819}}{0.789, 0.848} & \cint{\underline{0.808}}{0.780, 0.834} & \cint{\underline{0.805}}{0.779, 0.831} & \cint{\underline{0.800}}{0.774, 0.827} \\
& & \xmark & \cmark &
    \cint{0.920}{0.881, 0.950} & \cint{0.882}{0.849, 0.909} & \cint{0.838}{0.806, 0.866} & \cint{0.814}{0.786, 0.841} & \cint{0.800}{0.773, 0.827} & \cint{0.801}{0.774, 0.826} & \cint{0.794}{0.769, 0.819} \\
\rowcolor{gray!10}
& & \xmark & \xmark &
    \cint{0.923}{0.884, 0.953} & \cint{\underline{0.887}}{0.852, 0.917} & \cint{\underline{0.844}}{0.810, 0.876} & \cint{\textbf{0.820}}{0.790, 0.848} & \cint{\textbf{0.809}}{0.782, 0.837} & \cint{\textbf{0.809}}{0.783, 0.835} & \cint{\textbf{0.802}}{0.776, 0.828} \\
\midrule
\rowcolor{blue!15}
\multicolumn{11}{c}{\textbf{\textsf{M3FM Test Split}}} \\
\midrule
\multirow{1}{*}{Sybil}  & \multirow{1}{*}{SM-ST}
& \cmark & \cmark &
    \cint{0.943}{0.921, 0.962} & \cint{0.880}{0.846, 0.909} & \cint{0.847}{0.814, 0.876} & \cint{0.850}{0.818, 0.876} & \cint{0.849}{0.822, 0.874} & \cint{0.847}{0.821, 0.871} & \cint{0.844}{0.819, 0.869} \\
\midrule
\multirow{3}{*}{M3FM} & Large, MM-MT
& \xmark & \xmark &
    \cint{0.907}{0.874, 0.934} & \cint{0.855}{0.824, 0.883} & \cint{0.832}{0.804, 0.859} & \cint{0.814}{0.787, 0.840} & \cint{0.803}{0.776, 0.827} & \cint{0.803}{0.777, 0.828} & \cint{0.797}{0.771, 0.822} \\
& Huge, SM-ST & \xmark & \xmark &
    \cint{0.930}{0.898, 0.962} & \cint{0.870}{0.836, 0.904} & \cint{0.842}{0.809, 0.875} & \cint{0.834}{0.802, 0.865} & \cint{0.809}{0.777, 0.840} & \cint{0.792}{0.761, 0.824} & \cint{\text{--}}{\text{--}} \\
& Huge, MM-MT & \xmark & \xmark &
    \cint{0.940}{0.912, 0.970} & \cint{0.888}{0.857, 0.920} & \cint{0.860}{0.829, 0.891} & \cint{0.860}{0.831, 0.890} & \cint{0.839}{0.810, 0.869} & \cint{0.823}{0.794, 0.853} & \cint{\text{--}}{\text{--}} \\
\midrule
\multirow{3}{*}{\makecell{LungEvaty \\ (Ours)}} & \multirow{3}{*}{SM-ST}
& \cmark & \cmark &
    \cint{\textbf{0.947}}{0.923, 0.967} & \cint{\textbf{0.912}}{0.885, 0.934} & \cint{\textbf{0.898}}{0.873, 0.920} & \cint{\textbf{0.884}}{0.859, 0.908} & \cint{\underline{0.877}}{0.852, 0.897} & \cint{\underline{0.876}}{0.852, 0.895} & \cint{\underline{0.872}}{0.848, 0.892} \\
& & \xmark & \cmark &
    \cint{0.938}{0.913, 0.959} & \cint{0.899}{0.872, 0.922} & \cint{0.887}{0.860, 0.911} & \cint{0.879}{0.854, 0.902} & \cint{0.870}{0.847, 0.891} & \cint{0.870}{0.847, 0.889} & \cint{0.864}{0.842, 0.885} \\
\rowcolor{gray!10}
& & \xmark & \xmark &
    \cint{0.939}{0.912, 0.961} & \cint{\underline{0.901}}{0.873, 0.926} & \cint{\underline{0.891}}{0.864, 0.914} & \cint{\underline{0.883}}{0.860, 0.906} & \cint{\textbf{0.878}}{0.855, 0.898} & \cint{\textbf{0.878}}{0.855, 0.898} & \cint{\textbf{0.873}}{0.850, 0.893} \\

\bottomrule
\end{tabular}
\end{table*}

\begin{table}[!t]
\centering
\footnotesize
\setlength{\tabcolsep}{3pt}
\renewcommand{\arraystretch}{1.2} 
\caption{
\textbf{PR-AUC (95\% CI)} results on Sybil split, a more clinically relevant metric for this
imbalanced task. Our LungEvaty model (AIAG-free)
is compared to the best-performing Sybil (with Expert Anno.) and M3FM-Large (public model, as PR-AUC for Huge is not reported)
variants. Our model demonstrates performance gains across all 6 years.
}

\label{tab:prauc}

\begin{tabular}{l  *{6}{c}} 
\toprule
\multirow{2}{*}{\textbf{Model}}  & \multicolumn{6}{c}{\textbf{PR-AUC (Years)}} \\
\cmidrule(lr){2-7} 
 & Y1 & Y2 & Y3 & Y4 & Y5 & Y6 \\
\midrule
Sybil &
    \cint{0.357}{0.245, 0.473} & \cint{0.285}{0.203, 0.378} & \cint{0.251}{0.187, 0.330} & \cint{0.235}{0.178, 0.301} & \cint{0.244}{0.190, 0.303} & \cint{0.292}{0.238, 0.349} \\
\midrule

M3FM  &
    \cint{0.212}{0.138, 0.295} & \cint{0.179}{0.123, 0.245} & \cint{0.162}{0.115, 0.216} & \cint{0.151}{0.114, 0.198} & \cint{0.160}{0.126, 0.201} & \cint{0.220}{0.177, 0.266} \\
\midrule
\rowcolor{gray!10} 
\makecell{Lung-\\Evaty}

&
    \cint{\textbf{0.444}}{0.333, 0.547} & \cint{\textbf{0.364}}{0.273, 0.453} & \cint{\textbf{0.323}}{0.251, 0.397} & \cint{\textbf{0.297}}{0.231, 0.361} & \cint{\textbf{0.299}}{0.243, 0.356} & \cint{\textbf{0.361}}{0.306, 0.420} \\
\bottomrule
\end{tabular}
\end{table}

\section{Discussion and Conclusion}
Our results suggest that LungEvaty is a competitive framework for long-term risk prediction, particularly on the PR-AUC, a metric we believe is more clinically suitable for imbalanced screening data. As population-wide screening programs expand, the scalability of risk models becomes a crucial factor. Our design addresses this via a key trade-off: in exchange  for a pre-training phase and more parameters than Sybil \cite{mikhael2023sybil}, our model achieves superior ROC-AUC and PR-AUC performance without relying on pixel-level expert annotations, making it easier to apply to new data. Furthermore, our framework remains flexible and can incorporate the optional AIAG loss for additional spatial guidance when it is available. 
Compared to M3FM, we opted for a holistic, whole-lung approach, which we believe provides a more natural anatomical context. 
To preserve fine-grained detail within this large volume, we made the deliberate trade-off of using a small patch size of 8. While this choice increases sequence length and computational cost, it allows the model to learn both broad lung context and fine details simultaneously. We posit that this unified representation 
offers a more direct foundation for future extensions such as longitudinal risk prediction and multimodal learning that combines imaging with clinical text.

 
Our current evaluation is limited to the internal NLST test set. External validation on independent cohorts will be essential to confirm that the observed performance gains generalize across scanners, populations, and acquisition protocols. Moreover, extending these representations to other tasks and datasets will help verify the quality and versatility of our encoder. Assessing its performance on longitudinal data will clarify whether the learned features can capture disease progression. In addition, using the learned image representations for contrastive pretraining with text and tabular data could unlock multimodally pre-trained foundation models that link imaging with richer clinical information.

\clearpage
\section{Acknowledgements}
This project is supported by the Innovative Health Initiative Joint Undertaking (IHI JU) under grant
agreement No 101112135. The JU receives support from the European Union’s Horizon Europe
research and innovation programme and COCIR, EFPIA, Europa Bío, MedTech Europe, and Vaccines
Europe.
MC and JB are supported under grant agreement GA101080997.
RB received funding from the Bavarian State Ministry of Economic Affairs, Regional Department, and Energy as part of the Bavarian Joint Funding Program (BayVFP grant number DIK0444/03).
The authors thank the National Cancer Institute for access to NCI's data collected by the National Lung Screening Trial (NLST) (ClinicalTrials.gov identifier: NCT00047385, CDAS: NLST-1075). Special thanks to Peter Mikhael for their invaluable support in this work.

\section{Conflict of Interest}
The authors declare no conflict of interest.

\section{Compliance with Ethical Standards}
This study used de-identified data from the publicly available National Lung Screening Trial (NLST) datasets \cite{national2011national}. The NLST license confirms that no additional ethical approval is required for secondary analyses. As this work involved only retrospective analysis of de-identified data, no further institutional review was needed.

\bibliographystyle{IEEEbib}
\bibliography{strings,refs}


\end{document}